\documentclass[10pt,twocolumn,letterpaper]{article}

\usepackage{iccp}
\usepackage{times}
\usepackage{amsmath}
\usepackage{amssymb}
\usepackage{url}
\usepackage{graphicx}
\usepackage{epstopdf}

\iccpfinalcopy 

\def\iccpPaperID{11} 

\ificcpfinal\fi
\begin{document}

\title{Frequency Domain TOF: Encoding Object Depth in Modulation Frequency} 

\author{Achuta Kadambi\\
MIT Media Lab\\
{\tt\small achoo@mit.edu}
\and
Vage Taamazyan\\ 
Skoltech\\
{\small\url{vaheta@mit.edu}}
\and
Suren Jayasuriya\\
Cornell University\\
{\small\url{sj498@cornell.edu}}
\and 
Ramesh Raskar\\
MIT Media Lab\\
{\small\url{raskar@mit.edu}}
}

\maketitle
\thispagestyle{empty}

\begin{abstract}
Time of flight cameras may emerge as the 3-D sensor of choice. Today, time of flight sensors use phase-based sampling, where the phase delay between emitted and received, high-frequency signals encodes distance. In this paper, we present a new time of flight architecture that relies only on frequency---we refer to this technique as frequency-domain time of flight (FD-TOF). Inspired by optical coherence tomography (OCT), FD-TOF excels when frequency bandwidth is high. With the increasing frequency of TOF sensors, new challenges to time of flight sensing continue to emerge. At high frequencies, FD-TOF offers several potential benefits over phase-based time of flight methods.
\end{abstract}

\section{Introduction} 3D cameras are designed to capture depth of objects over a spatial field. The recent emergence of full-framerate 3D cameras such as the Microsoft Kinect have unlocked a number of applications in computer vision, computer graphics, and beyond. New applications for 3D cameras continue to surface, which places pressure on the demand for faster, more accurate, more flexible 3D systems that can operate in the wild. 

Today, there exist a large number of technologies which can acquire 3D information. Ongoing research efforts in depth sensing tend to exist as independent cells, e.g., theory from time of flight, which relies on timing optical echoes is largely disparate from that of stereo vision, which uses view-dependent parallax. In this paper we bring together two depth sensing technologies: time of flight cameras and optical coherence tomography. The former is embedded in Kinect cameras and the latter is an interferometric technique used to see microscopic details in 3-D structure. 

In this paper we restrict our scope to phase-based, time of flight (TOF) cameras. In such devices an active light source strobes in a coded pattern and the optical signal that returns to the focal plane is characterized by a shift in phase, which corresponds to the depth of an object. Phase-based TOF forms the basis for several popular cameras in machine vision, including the Microsoft Kinect, Mesa Swissranger, and PMD Camboard Nano. 

In all current implementations of phase TOF sensors, the phase is required to measure depth. Therefore, accurate measurement of depth relies on the accurate estimation of phase. This is challenging, for instance, in environments of multipath interference. An interesting question is whether it is possible to use the same sensor to obtain depth without using the phase? We answer this question by generalizing the specific theory of phase based time of flight sensors to the broader domain of optical coherence tomography. 

While phase TOF sensors operate on meter-size scenes, optical coherence tomography (OCT) works on small scenes and can obtain micron resolution of a 3D surface. OCT is often used in high-quality imaging of the human retina to diagnose glaucoma. There are two types of OCT: time-domain (TD-OCT) and frequency domain (FD-OCT). It turns out that phase TOF cameras, such as the Kinect, are using the same theoretical machinery from TD-OCT. In this paper, we are the first to bring the theory of FD-OCT to time of flight cameras.  

There are several benefits of our proposed technique of frequency-domain time of flight (FD-TOF). To obtain high accuracy depth maps it is necessary to increase the modulation frequency of TOF cameras. However, increasing the modulation frequency introduces several new challenges for the current, phase-based TOF architecture including phase wrapping and requirements on the phase shift resolution. 

To overcome these challenges \textbf{we propose a frequency-based architecture} that draws inspiration from the domain of optical coherence tomography. This technique is notable in that it does not require sampling in phase can therefore be implemented on either a TOF sensor or a conventional CCD/CMOS camera.

\section{Related Work} \begin{figure*}
\includegraphics[width=\textwidth]{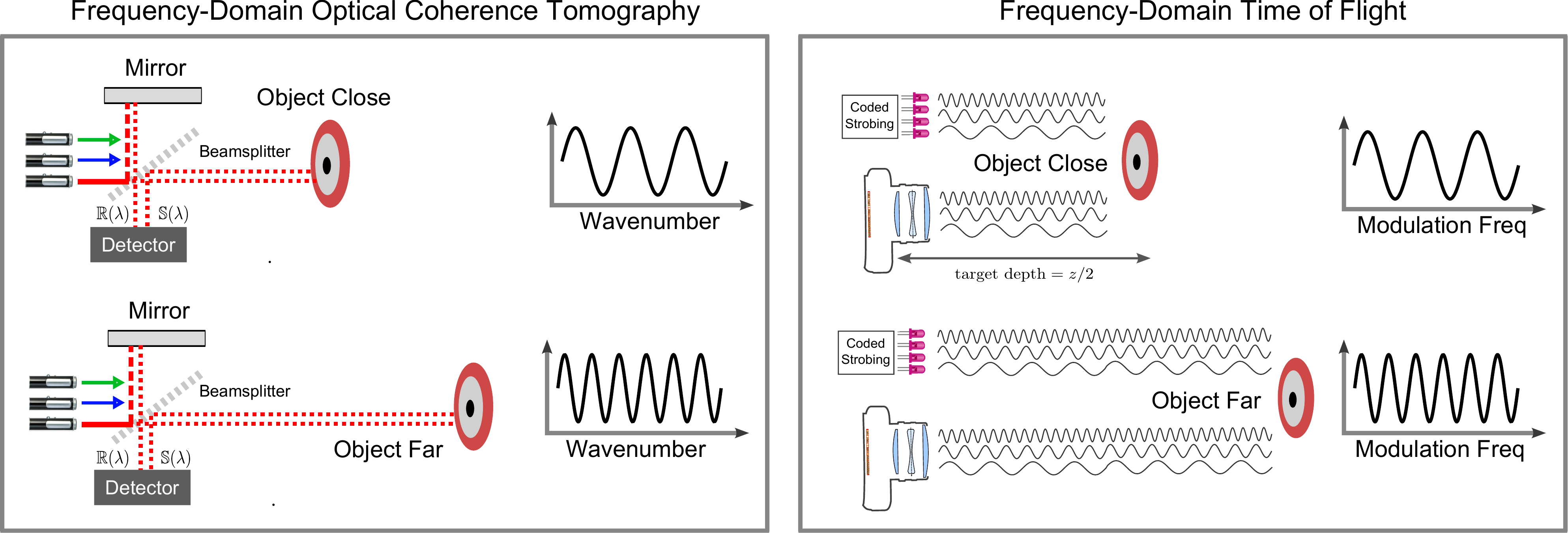}%
\caption{We introduce a new architecture for depth sensing: frequency-domain time of flight (FD-TOF). This concept is analogous to frequency-domain optical coherence tomography (FD-OCT). In particular, both techniques encode optical time of flight in the frequency of the received waveform. For short optical paths (top row), the received signal in the primal-domain is lower in frequency than that of longer optical paths (bottom row).}%
\label{fig:tofvsoct}%
\end{figure*}

Time of flight imaging is an emerging field in computational imaging. This paper strives to be self-contained: section \ref{ss:tofprelim} reviews time of flight, 3-D cameras from first principles. We use the term time of flight to refer to the time it takes a photon to travel a given distance through a medium. Work in computer vision and graphics centers around light transport analysis and algorithms as well as the capture of time of flight. 

\paragraph{Time of flight analysis} represents an increasingly popular area within the vision and graphics communities. For instance, Raskar and Davis \cite{raskar20085d} provided structure to the analysis of transient light transport by defining the 5-D light transport matrix. For a more complete overview of temporal light transport the reader is encouraged to review the work of Kutalakos \cite{seitz2005theory}, O’Toole \cite{o2014temporal}, Gupta \cite{gupta2014phasor} and Wu \cite{wu2014decomposing}. Time of flight analysis is closely linked to theory from tomography \cite{atcheson2008time,hullin2008fluorescent,ihrke2010transparent,alterman2014passive}, compressive imaging~\cite{ashok2011compressive,cevher2008compressive} and coded exposure imaging~\cite{hitomi2011video,veeraraghavan2011coded,agrawal2010optimal,asif2010streaming}. Combining transient light transport with such frameworks has spurred several applications, including ultra-fast imaging of light in flight \cite{velten2013femto,gariepysingle,heshmat2014single,heide2013low,kadambi2013coded}, scattered light imaging \cite{velten2012recovering,heide2014imaging}, measurement of BRDFs \cite{naik2011single}, new models for rendering \cite{jarabo2014framework}, and medical imaging \cite{satat2014locating}. A summary of space-time coding can be found in a recent report from Mitra et al.~\cite{mitra2013framework}. 

\paragraph{Time of flight capture} refers to the design and implementation of systems that can capture the time that a photon is in flight. Such devices always include an active light source and the goal is to measure the delay between light emission and detection. As the speed of light is fast---light travels one foot in a nanosecond---this seems like a challenging task. Fortunately, advances in opto-electronics have enabled wide accessibility to \textbf{phase TOF} cameras like the Microsoft Kinect. These cameras operate by strobing a modulated light source with a reasonable pattern, like a high-frequency square or sine wave. The signal that returns to the camera exhibits a depth-dependent phase shift, which is measured using time-correlation. In this paper, we break from the phase TOF paradigm to obtain depth using frequency alone. 

\paragraph{Multifrequency time of flight cameras} offer an additional set of measurements that have been used to improve the operation of phase TOF cameras. Recent work has used multifrequency measurements to address global illumination (cf. \cite{linfourier,payne2009multiple,bhandari2014resolving,heide2013low}), perform phase unwrapping or demultiplex illumination~\cite{kadambidemultiplexing}. Throughout these papers, the same phase TOF architecture is present, i.e., the received signal is time-correlated with a reference. In this paper we propose a new architecture that receives samples only in the frequency domain. This technique offers a new way to overcome challenges such as global illumination or phase unwrapping. 

\paragraph{Optical coherence tomography} is an optical interferometric technique that can obtain 3-D structures at micron resolution \cite{huang1991optical}. These devices are used extensively in biomedical imaging of structures such as the human retina. There are two main classifications: time domain OCT (TD-OCT) and frequency domain OCT (FD-OCT). The former is very similar to phase-based TOF cameras: the received signal is time-correlated with a reference signal to determine the phase offset. In contrast, frequency-domain OCT does not time correlate the received signal. Instead, the 3-D shape is obtained only by illuminating the sample at multiple optical frequencies. In this paper, we transpose the ideas from FD-OCT to the realm of time of flight 3-D cameras. 

\newcommand{\depth}{z} 
\newcommand{\modf}{f_{\text{M}}}
\newcommand{\ambient}{\beta}

%

\section{Preliminaries}

We begin our discussion of time of flight from first principles. Specifically, we describe the basic principles of phase TOF cameras, such as the Microsoft Kinect (Section \ref{ss:tofprelim}). Then we provide a condensed overview of optical coherence tomography in Section \ref{ss:octprelim}. We omit details of OCT that are specific to the optics community and thus less relevant to time of flight imaging. For a much more in-depth overview of OCT, the reader is encouraged to consult~\cite{huang1991optical} and~\cite{fujimoto2006optical}. 

\paragraph{Terminology:} We use the term \emph{primal-domain} to refer to the original frame that a signal is sampled in. We use the term \emph{dual-domain} to refer to the frequency domain with respect to the primal. 

\paragraph{Note:} To simplify exposition, all equations are provided in the context of a single scene point. The equations are spatially invariant. 

\subsection{Phase TOF architecture}
\label{ss:tofprelim}
A \textbf{phase TOF} camera is able to measure the phase delay of optical paths and obtain depth 
through the relation 
\begin{equation}
\depth = \frac{c \varphi}{2 \pi \modf} \quad \quad d = z/2, 
\label{eq:phi2d}
\end{equation} 
where $z$ is the total path length of a reflection, $d$ is the depth, $f_\text{M}$ is the modulation frequency of the camera and $c$ is the speed of light. The modulation frequency of the camera is around $100$ MHz, which corresponds to a period of 10 nanoseconds. To estimate $\varphi$ with high precision,
a phase TOF camera contains an active illumination source that is strobed according to a periodic illumination 
signal. In standard implementations (e.g. MS Kinect) the emitted signal 
takes the form of a sinusoid
\begin{equation}
g(t) = \cos \left(\modf t \right), 
\end{equation} 
where to simplify notation, we assume the emitted signal has unit amplitude. At the sensor plane, the received optical signal can be written as
\begin{equation}
s(t) = \alpha \cos \left(\modf t + \varphi \right) + \beta, 
\label{eq:recvA} 
\end{equation} 
where $\alpha$ is the attenuation in the projected
amplitude and $\beta$ is the intensity of ambient light. It is assumed that parameters $\alpha$, $\varphi$, and $\beta$ are time invariant within the exposure time. To estimate these three parameters we would need to sample Equation \ref{eq:recvA} at least three times within the period. Such sampling in the time-domain is challenging as it requires short, nanosecond exposures. Instead a TOF camera computes the cross-correlation of the emitted and received signals: 
\begin{equation}
c(\tau) = s(t) \otimes g(t) = \frac{\alpha}{2} \cos \left( \modf \tau + \varphi \right) + \beta. 
\label{eq:xcorr} 
\end{equation} 
Now, the primal-domain has changed from time, to $\tau$. This plays a key role, as it is physically easier to sample $\tau$ at nanosecond timescales. Specifically, this is done by introducing a nanosecond buffer between emitted and reference signals. To recover the phase and amplitude from the received signal, TOF cameras capture $N$ samples in the primal-domain and, in software, compute an $N$-point discrete Fourier transform (DFT). Suppose that four evenly spaced samples are obtained over the length of a period, for instance, $\tau = [0, \frac{\pi}{2}, \pi, \frac{3 \pi}{2}]^{T}$. Then the calculated phase can be written as
\begin{equation}
{\varphi} = \arctan \left( \frac{ c(\tau_4) - c(\tau_2) }
{ c(\tau_1) - c(\tau_3)} \right ),
\label{eq:4phi}
\end{equation}
and the calculated amplitude as 
\begin{equation}
\alpha  = \frac{1}{2}\sqrt {{{\left( {c\left( {{\tau _4}} \right) - c\left( {{\tau _2}} \right)} \right)}^2} - {{\left( {c\left( {{\tau _1}} \right) - c\left( {{\tau _3}} \right)} \right)}^2}}. 
\label{eq:4alpha}
\end{equation}
The two real quantities of amplitude and phase can be compactly represented as a single complex number using phasor notation: 
\begin{equation}
\mathbb{M} = \alpha e^{j \varphi},  
\end{equation} 
where $\mathbb{M} \in \mathcal{C}$ is the measured phasor, and $j$ is the imaginary unit. Armed with the phase, a TOF sensor computes depth using Equation \ref{eq:phi2d} and provides a measure of confidence using the amplitude. This concludes our overview of the standard phase TOF operation. 

\paragraph{Multipath interference:} We now describe the common multipath interference (MPI) artifact that affects phase TOF sensors. In such cases, $K$ reflections return to the imaging sensor and the received signal can be written as
\begin{equation}
c_{\text{MP}}(\tau) = \frac{1}{2} \left( \sum\limits_{l=1}^{K}{ \alpha_l \cos \left( \modf \tau + \varphi_l \right)} \right) + \ambient, 
\label{eq:mpxcorr} 
\end{equation} 
where the subscript $\text{MP}$ denotes multipath corrupted measurements. The received signal is now a summation of sinusoids at the same frequency but different phases. Obtaining the direct bounce, i.e., $K=1$, is a very challenging problem. After simplification using Euler's identity, the measured amplitude and phase in the presence of interference can be written as
\begin{equation}
\varphi_{\text{MP}} = \arctan \left( \frac{\sum_{i=1}^{K} \alpha_i \sin \varphi_i}{\sum_{i=1}^{K} \alpha_i \cos \varphi_i} \right) 
\end{equation}
\begin{equation}
\alpha_{\text{MP}}^2 = \sum\limits_{i=1}^{K} \alpha_{i}^{2} + 2 \underbrace{\sum\limits_{i=1}^{K} \sum\limits_{j=1}^{K} \alpha_i \alpha_j \cos \left( \varphi_i - \varphi_j \right)}_{i \neq j}. 
\end{equation}
Extracting the direct bounce from these corrupted phase and amplitude measurements turns out to be a challenging, non-linear inverse problem. The community has no clear solution, as all techniques to date are ill-conditioned~\cite{bhandari2014resolving,freedman2014sra}. Later in this paper, we return to the multipath problem, which allows separation of optical paths when using a frequency-domain architecture.

\paragraph{Increasing the modulation frequency:} Increasing the modulation frequency of TOF sensors allows for greater depth precision~\cite{lange2001solid}. Intuitively, at high frequencies, a small change in the estimated phase, corresponds to a small change in the estimated depth. Specifically, we can write the range accuracy $\Delta L$ as proportional to 
\begin{equation}
\Delta L \propto c/\modf. 
\end{equation}
Over the past few years, time of flight cameras have sported increased modulation frequencies to boost depth precision. For instance, the new Kinect tripled the modulation frequency over its competitors, increasing the modulation frequency to about 100 MHz. However, this frequency is at the upper limit of what the phase TOF architecture can handle. In particular, phase wrapping will occur for scene objects at a depth greater than
\begin{equation}
d_{\text{ambiguity}} = c / 2 \modf. 
\label{eq:pwrap}
\end{equation}
Unwrapping, or disambiguation of phase, requires more measurements in combination with a lookup table and is susceptible to noise~\cite{kolb2009time}. Another challenge that emerges at high frequencies is an increase in the required sampling rate in the primal-domain. 
\\
\\
\noindent \emph{Example:} Suppose the modulation frequency is $1$ GHz, which corresponds to a period of $1$ nanosecond. Two challenges arise. First, from Equation \ref{eq:pwrap}, objects further than 13 centimeters will cause phase ambiguities. Second, since $N$ samples are computed within the period, this means that $\tau$ has to be sampled at sub-nanosecond timescales, which is challenging to implement electronically. While using high modulation frequencies with phase TOF introduces new challenges, our proposed architecture is geared toward high modulation frequencies. 


\begin{figure*}
\includegraphics[width=\textwidth]{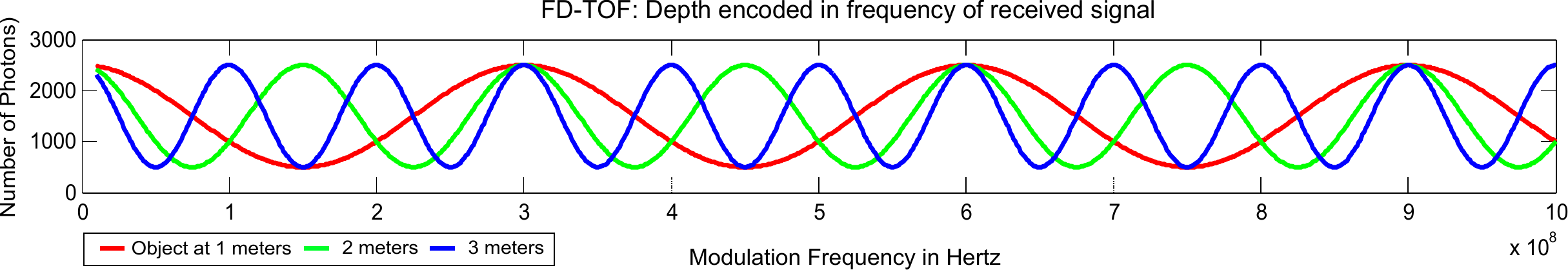}%
\caption{FD-TOF recasts depth estimation as a frequency estimation problem. The received signal is plotted in primal-domain for three different depths. The closest object generates the lowest frequency signal in primal-domain, while the furthest object generates the highest.}
\label{fig:fdtofbasic}%
\end{figure*}

\subsection{Primer on Optical Coherence Tomography} 
\label{ss:octprelim}
While phase TOF is a type of electronic interferometry, optical coherence tomography performs interferometry directly on the optical signal. Specifically, depth is obtained with respect to a reference sample by correlating reflections from the reference with the sample.  OCT is divided into two factions, which are characterized by the type of sampling they use. The older, Time-Domain OCT (TD-OCT) technique uses time-correlations of an optical reference with a sample (to correlate in time, the reference sample is shifted). This is precisely the mechanism of phase TOF described in section \ref{ss:tofprelim}. The second class of technique is Frequency-Domain OCT (FD-OCT), which performs sampling only in frequency domain. In this section, we provide a concise overview of FD-OCT, describing only the facets that can be applied to 3D cameras. 

In Frequency-Domain OCT, depth is obtained by sampling the signal at different optical wavelengths (i.e. wavelength is the primal-domain). Figure \ref{fig:tofvsoct} provides a schematic for the typical FD-OCT system. At a single wavelength, the detector receives an electric field from the reference object, which takes the form of 
\begin{equation}
\mathbb{R}\left(\lambda \right) = \alpha(\lambda){e^{j\varphi_R \left(\lambda \right)}}, 
\end{equation} 
where $\mathbb{R}(\lambda)$ represents the received phasor as a function of optical wavelength. 
Similarly, the received electric field from the sample object is written as
\begin{equation}
\mathbb{S}\left(\lambda \right) = \alpha(\lambda){e^{j\varphi_S \left(\lambda \right)}}. 
\end{equation}
Note that the amplitude of the sample and reference are assumed to be equal, which simplifies our explanation of the concept. A combination of the two reflections return to an imaging sensor. The electric field at the detector is the summation
\begin{equation}
\mathbb{M}\left(\lambda \right) = \frac{1}{2} \left( \mathbb{R} \left(\lambda \right) + \mathbb{S} \left(\lambda \right) \right), 
\label{eq:m} 
\end{equation}
where it is assumed that the constituent phasors are halved when they recombine (for instance, due to a beamsplitter). The current measured at the detector can be expressed as the real quantity
\begin{equation}
i \left(\lambda \right) = \frac{\eta q}{h \nu} \left| \mathbb{M} \left(\lambda \right) \right|^2,  
\label{eq:i} 
\end{equation}
where $\eta$ is the detector sensitivity, $q$ is the quantum of electric charge, $h$ is Planck's constant, and $\nu$ is the optical frequency. By substituting Equation \ref{eq:m} into \ref{eq:i} we obtain
\begin{equation}
\resizebox{1\hsize}{!}{$
i \left(\lambda \right) = \frac{1}{4} \frac{\eta q}{h \nu} \left( \underbrace{\mathbb{R}\left(\lambda \right) \left(\mathbb{R} \left( \lambda \right) \right)^{*} + \mathbb{S}\left(\lambda \right) \left(\mathbb{S} \left( \lambda \right) \right)^{*} }_{\text{Autocorrelation}} + \underbrace{2 Re \left( \left(\mathbb{R} \left( \lambda \right) \right)^{*} \mathbb{S}\left(\lambda \right) e^{-j \varphi_z} \right)}_{\text{Crosscorrelation}} \right). 
$}
\label{eq:bigi}
\end{equation}
Here, $\varphi_{z}$ represents the phase delay due to the difference in optical path length between the reference and optical reflections. Similar to the TOF case, phase and z-distance are related:
\begin{equation}
\varphi_z = 2 \pi z / \lambda.  
\label{eq:z2phi2}
\end{equation}
In Equation \ref{eq:bigi} note that the Autocorrelation terms are DC with respect to the wavelength. By using this relation along with Equation \ref{eq:z2phi2}, we can rewrite Equation \ref{eq:bigi} as
\begin{equation}
i \left(\lambda \right) = \frac{1}{2} \frac{\eta q}{h \nu} \left( \alpha \left(\lambda \right) \right)^2 \left( 1 + 2 \cos \left( \frac{2 \pi z}{\lambda} \right) \right). 
\label{eq:simplei}
\end{equation}
Now we introduce an auxiliary variable $k = 2 \pi / \lambda$, which is known as the wave number. Equation \ref{eq:simplei} can be rewritten as
\begin{equation}
i \left(k \right) = \frac{1}{2} \frac{\eta q}{h \nu} \left( \alpha \left(k \right) \right)^2 \left( 1 + 2 \cos \left( kz \right) \right), 
\label{eq:simplek}
\end{equation}
where now the primal-domain is the wavenumber ($k$). The dual can be computed as 
\begin{equation}
{\cal F}\left[ {i \left(k \right)} \right]\left( \kappa \right) \propto \delta \left( \kappa  \right) + \delta \left( {\kappa \pm z} \right),
\end{equation} 
where $\kappa$ represents the dual domain. To summarize: in frequency-domain OCT, reflections at multiple wavenumbers are measured at the detector and depth is encoded in the frequency of the received signal in primal-domain. 

\paragraph{Note on Depth Resolution:} The frequency bandwidth that FD-OCT can sample is quite large and is related to the difference between the largest and shortest wavelengths that are used to illuminate the sample. A large bandwidth allows for very fine axial, or depth resolution. Axial resolution is different from depth precision---it refers to the minimum separation between layers that can be resolved. For FD-OCT, the bound for axial resolution is complicated, and depends on the optical hardware. However, in section \ref{ss:tofres}, we show that a similar derivation exists for time of flight cameras and can be expressed in a simpler form.


\section{Frequency Domain Time of Flight} 

Inspired by Frequency-Domain OCT, we can reexamine the conventional operation of TOF cameras. In this section we provide a recipe for depth estimation by sampling different modulation frequencies at a single phase step. We refer to this new architecture as \textbf{FD-TOF}, where the primal-domain is modulation frequency.   

\subsection{Depth sensing using only modulation frequency} 

In this section we show how depth can be calculated, not from phase steps, but from frequency steps. Recall from Section \ref{ss:tofprelim} that the received signal takes the form of
\begin{equation}
c(\tau) = \frac{\alpha}{2} \cos \left( \modf \tau + \varphi \right) + \ambient. 
\end{equation} 
In standard phase TOF, $\tau$ represents the primal-domain against which we would ordinarily compute the $N$-point DFT. Instead, consider substituting 
Equation \ref{eq:phi2d} into Equation \ref{eq:xcorr} to obtain
\begin{equation}
c(\tau, \modf) = \frac{\alpha}{2}  \cos \left( \modf \tau + \frac{2 \pi z}{c} \modf \right) + \ambient. 
\end{equation} 
Without loss of generality assume that this signal is sampled only at the zero shift, i.e., $\tau = 0$. Then the received signal
at the sensor takes the form of
\begin{equation}
c(\tau = 0, \modf) = c(\modf) = \frac{\alpha}{2}  \cos \left( \frac{2 \pi z}{c} \modf \right) + \ambient. 
\end{equation} 
Now the primal domain is $\modf$ and the associated dual takes the form of
\begin{equation}
\mathcal{F}\left[c\left(\modf \right) \right]\left( \kappa \right) \propto \delta \left(\kappa \right) + \delta \left(\kappa \pm \frac{2 \pi z}{c} \right),  
\end{equation} 
where $\kappa$ is the dual-domain, corresponding to the inverse of the modulation frequency. Analogous to FD-OCT the depth can be obtained by finding the location of the support in the dual domain. 

\subsection{Multipath interference in FD-TOF} 

An advantage of FD-TOF is that multipath interference is separable in the dual-domain. Recall that in the multipath problem, $K$ reflections return to the sensor and the received signal is given by
\begin{equation} 
c(\modf) = \frac{1}{2} \left( \sum\limits_{l = 1}^K {\alpha_l \cos \left( {\frac{{2\pi {z_l}}}{c} \modf} \right)} \right)+ \ambient. 
\end{equation} 
The associated Fourier transform can now be written as
\begin{equation} 
\mathcal{F}\left[c\left(\modf \right) \right]\left( \kappa \right) \propto \delta \left(\kappa \right) + \sum\limits_{l=1}^K {\alpha_l \delta \left(\kappa \pm \frac{2 \pi z_l}{c} \right)}.
\label{eq:recv}
\end{equation}
Here, the received signal is a sum of sinusoids at the same phase but at different frequencies. Assume for the moment that the signal is sampled adequately to avoid aliasing. Then the different depths can be resolved by analyzing the location of spectral peaks in Fourier domain. For comparison in phase-TOF the received signal is a sum of sinusoids at the same frequency but different phases. Here, the inverse problem of recovering the constituent depths is ill-posed.

\begin{figure*}%
\centering
\includegraphics[width=\textwidth]{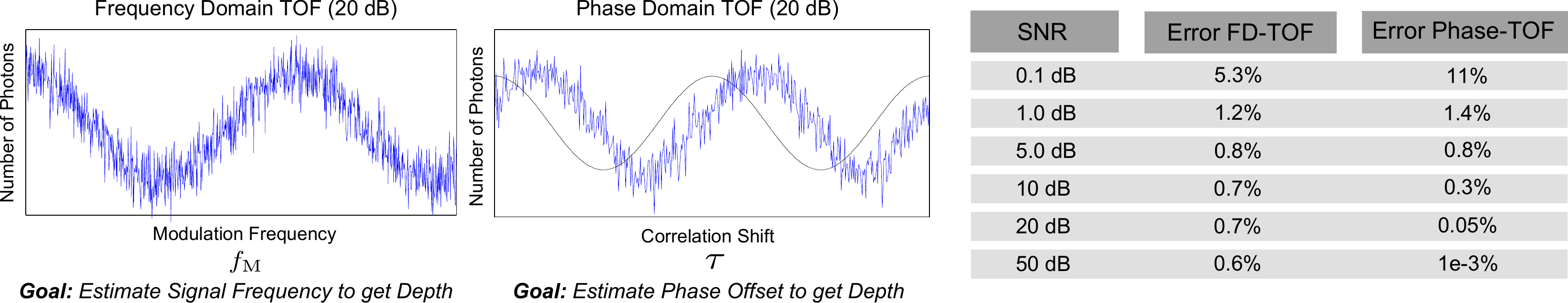}%
\caption{FD-TOF estimates frequency in the primal-domain of $\modf$, while phase TOF estimates phase in the primal-domain of $\tau$. At left, the primal-domain for FD-TOF and phase TOF at 20 dB SNR. The table at right lists percent error for different levels of SNR. FD-TOF performs better for depth recovery when the SNR is low and phase TOF performs better at high levels of SNR.}%
\label{fig:noise}%
\end{figure*}

\subsection{Estimating frequency to obtain depth} 

FD-TOF recasts depth estimation into the problem of frequency estimation in the received signal (Equation \ref{eq:recv}). Frequency estimation of a discrete waveform is a well-studied problem in the digital signal processing community. While Fourier analysis is simple and well-known, there are several, more statistically-efficient ways to estimate the frequency. For our results, we use Quinn and Fernandes technique~\cite{quinn1991fast}, which is as efficient as the least squares estimator of frequency. 

There are several other ways to estimate the frequency, including Pisarenko's method and Multiple Signal Classification (MUSIC). The reader is encouraged to see Stoica et al. \cite{stoica1997introduction} for an excellent overview of techniques. 

\subsection{Frequency bandwidth and resolution} 
\label{ss:tofres}

We now turn to the concept of axial or depth resolution. The concept is used in OCT to characterize the minimum distance at which two layers can be separated. We adapt this concept to TOF sensors to characterize multipath interference. In this context, the modulation frequencies that are sampled can be described as a boxcar function in the primal-domain
\begin{equation}
\Pi \left( \modf \right)  = H \left( \modf - \modf^{-} \right) - H \left( \modf - \modf^{+} \right), 
\end{equation}
where $\modf^{-}$ and $\modf^{+}$ represent the minimum and maximum modulation frequencies that are sampled and $H(\cdot)$ refers to the Heaviside step function. The Fourier transform of $\Pi \left( \modf \right)$ takes the form of a sinc function: 
\begin{equation}
\mathcal{F}\left[\Pi \left( \modf \right) \right]\left( \kappa \right) \propto \Delta \modf \frac{ \sin \Delta \modf \kappa}{\kappa}, 
\end{equation} 
where $\Delta \modf = \modf^{+} - \modf^{-}$. The FWHM of this function determines the axial resolution $\Delta z$, which is approximately:
\begin{equation}
\Delta z \approx 1.2c/\Delta \modf. 
\label{eq:mpibound}
\end{equation} 
In summary, if the frequency bandwidth $\Delta \modf$ is large, it is possible to disentangle multipath and scattering. If the frequency bandwidth becomes sufficiently high, FD-TOF may find relevance in biomedical imaging of small-scale structures. 
\\
\\
\noindent \emph{Example:} Current TOF sensors can support a frequency bandwidth of approximately 100 MHz. Using Equation \ref{eq:mpibound}, multipath interference can be disentangled in frequency domain if the optical paths differ by about 3.6 meters. If the frequency bandwidth increases by two orders of magnitude, then optical paths that differ by a few centimeters can be separated.    

\section{Slow-TOF: Can we use a normal camera?} 
\label{sec:dslr}

Sampling the primal-domain in FD-TOF amounts to changing the strobing frequency of the light source. Since only the light frequency is changing, a reasonable question is whether it is possible to use a conventional camera (hereafter ``slow camera''). We refer to this proposed technique as \textbf{Slow-TOF}. Recall that a slow camera integrates the photon flux over an exposure time. The image formation model in the primal domain is given by
\begin{equation}
I(\modf) = \int_{0}^{t_E} \alpha \cos \left( \modf t + \frac{2 \pi z}{c} \modf \right) + \beta\,dt, 
\end{equation}
where $t_E$ denotes the exposure time. After expanding the integral, the measured intensity can be written as a summation of two sinusoids: 
\begin{equation}
\resizebox{1\hsize}{!}{$
I(\modf) = \underbrace{\frac{\alpha}{\modf} \left[ \sin \left( \modf t_E + \frac{2 \pi z}{c} \modf \right) - \sin \left( \frac{2 \pi z}{c} \modf \right) \right]}_{\text{AC component}} + \ambient t_E. 
$}
\label{eq:dslr}
\end{equation} 
As expected, the AC component of Equation \ref{eq:dslr} evaluates to zero when the exposure time is zero. However, if the exposure time is greater than zero, then the measured image intensity is a summation of two sinusoids. The Fourier transform of Equation \ref{eq:dslr} with respect to $\modf$ is proportional to 
\begin{equation}
\resizebox{1\hsize}{!}{$
\mathcal{F}\left[c\left(\modf \right) \right]\left( \kappa \right) \propto \delta \left( \kappa \right) + \delta \left(\kappa \pm \left( \frac{2 \pi z}{c} + t_E \right) \right) + \delta \left(\kappa \pm \frac{2 \pi z}{c} \right). 
$}
\end{equation} 
As before, the depth can be calculated by finding the location of support in the dual domain. In practice, Slow-TOF introduces two additional challenges: 
\begin{itemize}
\item The AC component of the received signal decays at high modulation frequencies, and
\item The exposure time needs to be consistent during the sampling process. 
\end{itemize} 
The first point is clear from the form of Equation \ref{eq:dslr}, where the amplitude is inversely proportional to the frequency of modulation. One way to address this challenge is to increase the amplitude of the light source. For the second point, if the exposure time is changing while sampling in frequency, then the sampling will not be meaningful. For this reason, implementing Slow-TOF with a consumer-grade DSLR may be challenging as the exposure is not consistent. We believe a successful implementation of Slow-TOF would need to use a machine vision camera. 

\section{Experiments}
To validate that depth can indeed be encoded in frequency, we perform real and simulated experiments.

\begin{figure}
\includegraphics[width=\columnwidth]{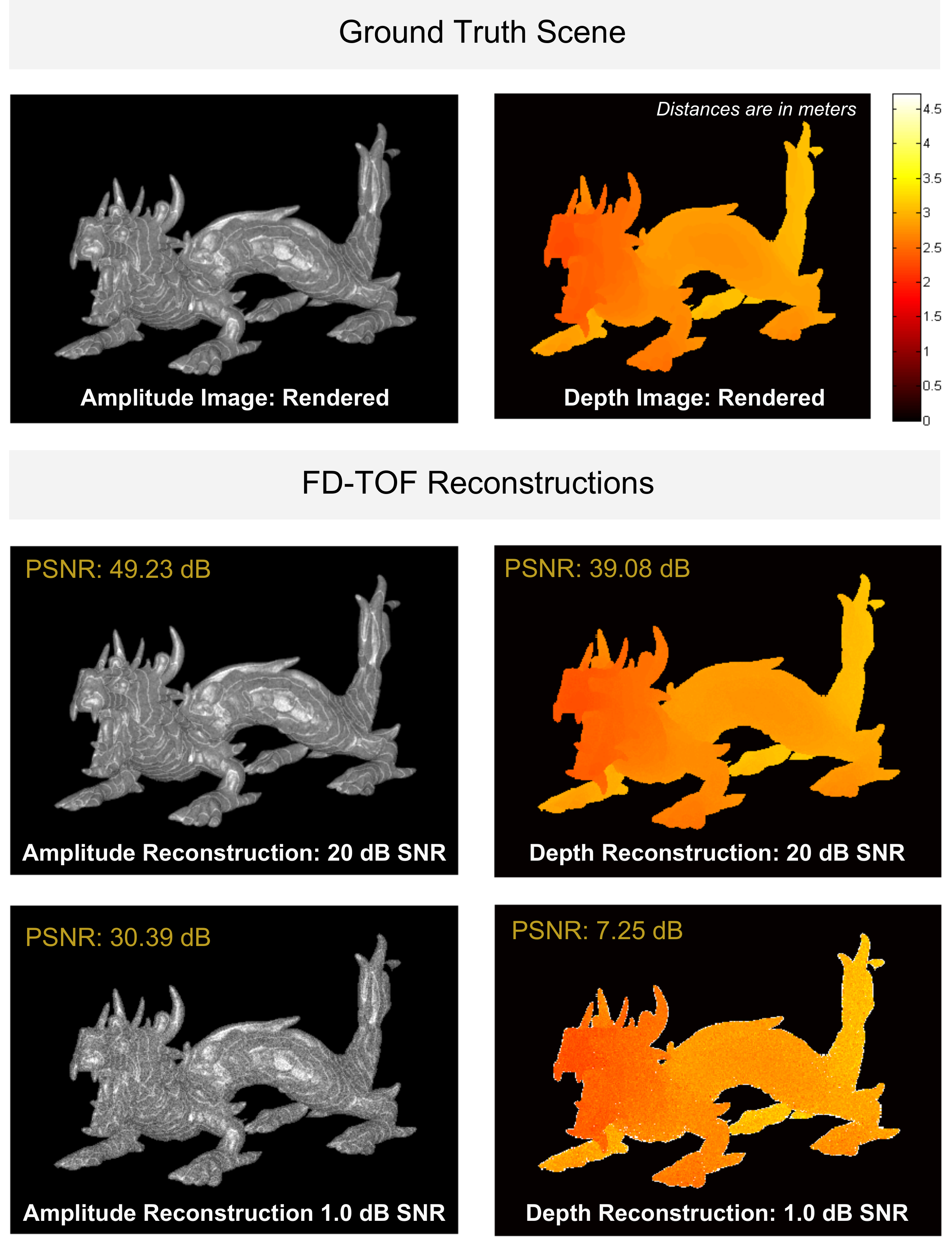}%
\caption{Recovery of complex scenes. Here, we render a scene with a dragon exhibiting scattering and interreflections. The first row shows ground truth amplitude and depth images (colorbar represents depth in meters). We are able to reconstruct the dragon at 20 dB and 1 dB SNR.}%
\label{fig:dragon}%
\end{figure}

\begin{figure}%
\includegraphics[width=\columnwidth]{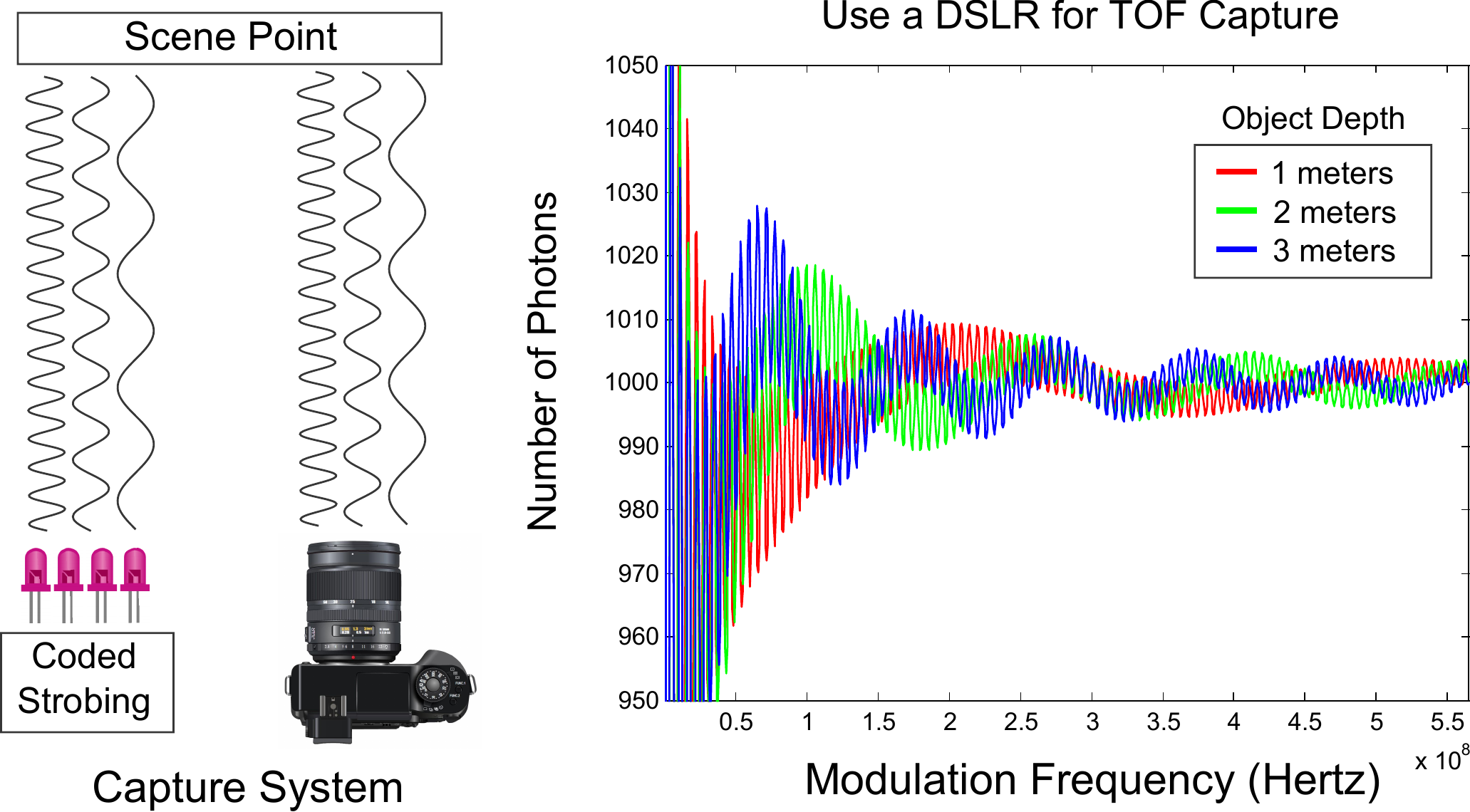}%
\caption{It is possible to use a conventional camera to recover time of flight. In FD-TOF the primal-domain is the modulation frequency. By strobing the illumination at different modulation frequencies, the received signal at the slow camera exhibits a depth-dependent frequency. In the plot at right, the furthest object has the highest frequency in the primal domain, while the closest object exhibits the lowest frequency. Note the decay in amplitude of the signal as the modulation frequency increases.}
\label{fig:dslr}%
\end{figure}

\subsection{Simulated Experiments} 

\paragraph{FD-TOF: Noiseless Simulation} To illustrate the basic concept we first consider a noiseless simulation. Figure \ref{fig:fdtofbasic} plots the received signal in the primal domain (i.e. modulation frequency) for three different scenes. The modulation frequency ranges from 10 MHz up to 1 GHz and the three curves correspond to a single object placed at 1, 2 and 3 meters. When the object is at the 1 meter position, the received signal is a low frequency sinusoid (red curve). In contrast, when the object is placed further away, e.g., at 2 or 3 meters, the received signal doubles or triples in frequency.  

\paragraph{FD-TOF: Performance with Noise} A key benefit of frequency estimation lies in its robustness to noise. If the received signal consists of a single tone, recovering the frequency of the sinusoid is a problem with robust guarantees. We compare our proposed, frequency domain TOF technique with the conventional phase domain TOF technique in the presence of noise. As illustrated in Figure \ref{fig:noise}, the goal of FD-TOF is to measure the frequency of the received, noisy signal (plotted at 20 dB SNR). For comparison, in phase domain TOF the primal-domain is $\tau$, the phase shift. The goal of this architectures is to measure the phase shift in the primal domain between a noisy, received signal with a reference signal. As illustrated in the table of Figure \ref{fig:noise}, the percent error of phase and frequency domain TOF is comparable at different levels of SNR. As a general trend, FD-TOF has more robust performance at very low levels of SNR, while phase domain TOF has improved performance at high levels of SNR. To summarize: both techniques are robust to noise, however FD-TOF performs better in cases of low SNR.

\paragraph{FD-TOF: Complex object} To analyze the performance of FD-TOF on a complex scene, we use the Mitsuba software~\cite{jakob2010mitsuba} to render a model of a Dragon. This scene consists of a Lambertian dragon with a small amount of scattering due to interreflections. The ground truth amplitude and depth images are illustrated in the first row of Figure \ref{fig:dragon} and reconstructions at 20 dB and 1 dB SNR occupy the second and third rows of the figure. 20 dB SNR is a plausible value for a real-world camera system. At this noise level the dragon is reconstructed accurately. Even at 1 dB SNR, the depth reconstruction has a PSNR of 7.25 dB. 

\paragraph{Slow-TOF: Simulations} Recall that in Section \ref{sec:dslr}, we suggested that FD-TOF can be implemented on a slow camera. As illustrated in Figure \ref{fig:dslr} the proposed architecture consists of the same coded, strobing that is used on TOF sensors. However, unlike a TOF sensor, a regular DSLR integrates the photon flux over an exposure time. This integrated value is plotted for different strobing frequencies in Figure \ref{fig:dslr}. Observe that the objects at $1$ , $2$ and $3$ meters have distinct curves in the primal-domain, which can be separated. Note that at higher modulation frequencies the signal amplitude decays (cf. Equation \ref{eq:dslr}). In order to glean information from the higher frequency bands it is necessary to use a light source with a large amplitude. 

It is interesting to note that, in the simulation it is possible to distinguish nanosecond delays of light travel (light travels 1 foot in a nanosecond). This result implies that with a slow camera like a DSLR, it is theoretically possible to capture nanosecond phenomena.

\begin{figure*}
\includegraphics[width=\textwidth]{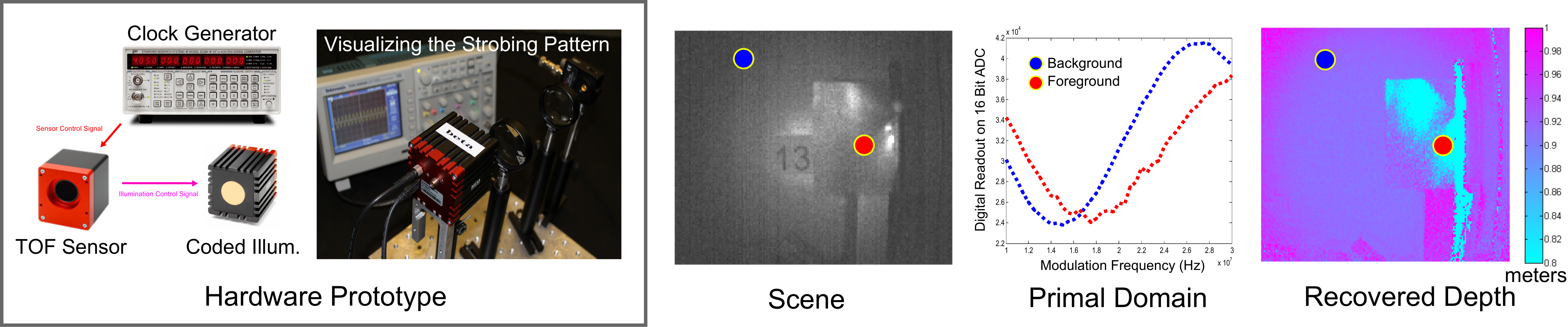}%
\caption{Validation of FD-TOF with a real experiment. At left is the hardware prototype---a TOF camera that can be customized to accept arbitrary clock signals. The oscilloscope shows the strobing pattern of the light source in time. The scene is shown in the center and consists of a wall at 1 meter and a sheet of paper at 80 cm. The blue point is further away and has a slightly higher frequency than the red point when plotted in the primal-domain.}%
\label{fig:real}%
\end{figure*}

\subsection{Physical Experiments}

\paragraph{Implementation:} To implement FD-TOF the frequency of the TOF sensor and light source must be sampled with high granularity. The Mesa Scientific Lock-in Module (SLIM) is available for purchase from Mesa Imaging in Zurich. This camera is essentially a bare TOF sensor that accepts a range of clock signals from 10 MHz to 30 MHz. We add a standard DSLR lens (Canon EF-S 18-55 mm), a clock generator (Stanford Research DS345) and a light source bank (twelve 850 nm LEDs). The clock generator provides the reference code to the camera, which in turn provides the strobing pattern to the light bank. By placing a photodiode in front of the strobing light source we are able to verify the coded strobing pattern on an oscilloscope. The hardware implementation is shown in Figure \ref{fig:real}. 

\paragraph{FD-TOF reduced to practice:} As illustrated in Figure \ref{fig:real}, the scene consists of a white wall approximately 1 meter away from the camera. For the foreground object, we place a sheet of paper approximately 80 centimeters away from the camera. In Figure \ref{fig:real} we plot the received signal in the primal-domain for a foreground and background pixel. When plotting each pixel in primal-domain, we observe that the further away pixel exhibits a higher frequency. Since the primal-domain ranges from only 10 MHz to 30 MHz, we observe barely one cycle in primal-domain, and the change in frequency is harder to detect. This factor, combined with the low signal from portions of paper affect the quality of the full depth reconstruction where part of the paper is not rendered correctly. Nevertheless, the implementation validates our core idea: that depth can be encoded, not in phase, but in frequency. 

\section{Discussion} 
In summary we present a new architecture for time of flight 3D imaging by changing the primal-domain.  Instead of relying on measuring phase shifts, we recast the entire problem in the frequency domain. As the modulation frequency of TOF sensors continues to increase, the benefits of frequency-domain OCT will start to become accessible to TOF sensing. At GHz or THz frequencies, our proposed architecture will obtain micron resolution.  

\subsection{Overview of Benefits}
\noindent The benefits of FD-TOF include
\begin{itemize}
\item Robustness to phase and multipath challenges,  
\item Compatibility with slow cameras, and 
\item A framework for flexible interferometry.  
\end{itemize}
 
\paragraph{Phase and multipath challenges:} Today, TOF sensors are at a critical point where phase wrapping is a real challenge. Consider the following: a $100$ MHz time of flight camera has an ambiguity distance of only $1.33$ meters, which necessitates phase unwrapping techniques. While there are benefits to increasing modulation frequency, it is not clear if phase TOF will be fully compatible (phase unwrapping algorithms are susceptible to noise). In addition, at high frequencies, sampling in phase needs to be very precise, down to picoseconds. Finally, in phase TOF, resolving multipath interference is a challenging, non-linear inverse problem. With sufficient frequency bandwidth, the proposed FD-TOF architecture is robust to such challenges.    

\paragraph{Slow cameras can measure TOF:} We have shown that, in theory, time of flight can now be captured with a conventional, slow camera. This is enabled through the architecture of FD-TOF, where sampling in the primal-domain amounts to changing the frequency of the light source. Our analysis suggests that a slow camera, equipped with \emph{millisecond} exposures, has the potential to capture time of flight to \emph{nanosecond} precision. We believe that the implementation of such a device would lead to interesting follow-up work. 

\paragraph{Flexible Interferometry:} In this paper we establish a duality between electronic and optical interferometry. The former technique performs correlation on the \emph{modulated} signal, while the latter correlates the carrier signal. In some cases, electronic interferometry might be preferable to optical inteferometry. For instance, today, it is challenging to perform optical interferometry at GHz or THz frequencies due to well-known optical challenges (see "Terahertz Gap"). The architecture we have proposed is a step toward increasing the flexibility of interferometry. 

\subsection{Limitations} 
The proposed technique is not well-suited to available TOF hardware. First, current TOF sensors are designed to sample in phase, not in modulation frequency. Second, today's sensors only have about 100 MHz of frequency bandwidth, which means that not enough cycles of the sinusoidal signal will be observed in the primal-domain. In the absence of sufficient cycles, frequency estimation becomes a more challenging problem. Finally, we note that accurate tone estimation requires sampling a number of frequencies. Although in theory, a minimum of $3$ frequencies are required, in the presence of noise more may be required. Follow-up work would explore tradeoffs in frequency sampling (e.g., how many frequencies are required, which frequencies are optimal).  
 
\section{Conclusion}

We have demonstrated FD-TOF, a new architecture for TOF sensors. As modulation frequencies continue to increase, phase TOF cameras face new challenges, such as phase wrapping and phase stepping. Our proposed system may represent a step toward high-frequency, electronic interferometry.

{\small
\bibliographystyle{ieee}
\bibliography{iccp_references}
}

\end{document}